%% file: main.tex
\documentclass[10pt,twocolumn,letterpaper]{article}

\usepackage[pagenumbers]{cvpr} % To force page numbers, e.g. for an arXiv version

\input{preamble}

\definecolor{cvprblue}{rgb}{0.21,0.49,0.74}
\usepackage[pagebackref,breaklinks,colorlinks,citecolor=cvprblue]{hyperref}
\usepackage{graphicx}
\usepackage{amsmath}
\usepackage{amssymb}
\usepackage{booktabs}
\usepackage{amsmath,amsthm,amssymb,amsfonts}
\usepackage[ruled]{algorithm2e} %带竖线
\usepackage{enumitem} 
\usepackage{caption}
\usepackage{pifont}
\usepackage{multirow}
\usepackage{enumerate}
\usepackage{color}

\usepackage[pagebackref,breaklinks,colorlinks]{hyperref}

\newcommand{\Imat}[0]{\ensuremath{{\bf I}} }

\newcommand{\Tmat}[0]{\ensuremath{{\bf T}} }

\newcommand{\fv}[0]{\ensuremath{\boldsymbol{f}} }

\newcommand{\mv}[0]{\ensuremath{\boldsymbol{m}} }

\newcommand{\alphav}[0]{\ensuremath{\boldsymbol{\alpha}} }
\newcommand{\betav}[0]{\ensuremath{\boldsymbol{\beta}} }
\newcommand{\gammav}[0]{\ensuremath{\boldsymbol{\gamma}} }

\def\gg{\textcolor{gray}}

\usepackage[capitalize]{cleveref}
\crefname{section}{Sec.}{Secs.}
\Crefname{section}{Section}{Sections}
\Crefname{table}{Table}{Tables}
\crefname{table}{Tab.}{Tabs.}
\def\paperID{10884} % *** Enter the Paper ID here
\def\confName{CVPR}
\def\confYear{2024}

\title{MeaCap: Memory-Augmented Zero-shot Image Captioning}

\author{Zequn Zeng\thanks{Equal contribution. \hspace{4mm}  \textdagger Corresponding authors}, Yan Xie\footnotemark[1], Hao Zhang\footnotemark[2], Chiyu Chen, Bo Chen\footnotemark[2]\\
National Key Laboratory of Radar Signal Processing, Xidian University, Xi’an, 710071, China\\
{\tt\small\{zzequn99, yanxie0904, zhanghao\_xidian\}@163.com, \{chenchiyu, bchen\}@mail.xidian.edu.cn}
\and
 Zhengjue Wang\\
State Key Laboratory of Integrated Service Networks, Xidian University, Xi’an, 710071, China\\
{\tt\small wangzhengjue@xidian.edu.cn}
}

\begin{document}
\maketitle
\begin{abstract}
Zero-shot image captioning (IC) without well-paired image-text data can be divided into two categories, training-free and text-only-training.
Generally, these two types of methods realize zero-shot IC by integrating pre-trained vision-language models like CLIP for image-text similarity evaluation and a pre-trained language model (LM) for caption generation. 
The main difference between them is whether using a textual corpus to train the LM.
Though achieving attractive performance \wrt some metrics, existing methods often exhibit some common drawbacks.
%in these two categories have their own problems as follows.
Training-free methods tend to produce hallucinations, while text-only-training often lose generalization capability.
To move forward, in this paper, we propose a novel \textbf{Me}mory-\textbf{A}ugmented zero-shot image \textbf{Cap}tioning framework (\textbf{MeaCap}).
Specifically, equipped with a textual memory, we introduce a retrieve-then-filter module to get key concepts that are highly related to the image.
By deploying our proposed memory-augmented visual-related fusion score in a keywords-to-sentence LM, MeaCap can generate concept-centered captions that keep high consistency with the image with fewer hallucinations and more world-knowledge.
The framework of MeaCap achieves the state-of-the-art performance on a series of zero-shot IC settings.
Our code is available at
\href{https://github.com/joeyz0z/MeaCap}{https://github.com/joeyz0z/MeaCap}.

\end{abstract}
\vspace{-2mm}
\section{Introduction}
Image captioning (IC) aims to understand visual content and generate text descriptions.
Using well-annotated image-text pairs,  supervised models \cite {vaswani2017attention, kim2021vilt, liu2021swin, cornia2020meshed,huang2019attention, nguyen2022grit, pan2020x, schwartz2019simple, schwartz2017high} have achieved promising results on typical IC benchmarks \cite{lin2014microsoft,young2014image,krishna2017visual,agrawal2019nocaps}.
Due to the high costs of annotation, the training sets of these benchmarks often involve limited image styles/contents, which is a hard obstacle for those supervised models to be generalized to images in the wild.
To realize IC without human-annotated image-text pairs, recently, zero-shot IC has drawn increasing attention.
Existing works can be mainly divided into two groups, training-free methods and text-only-training methods. 
\begin{figure}[!t]
  %\label{fig:fig1}
  \centering
  \subfloat[Hallucination phenomenon.]{\includegraphics[width=1.0\linewidth]{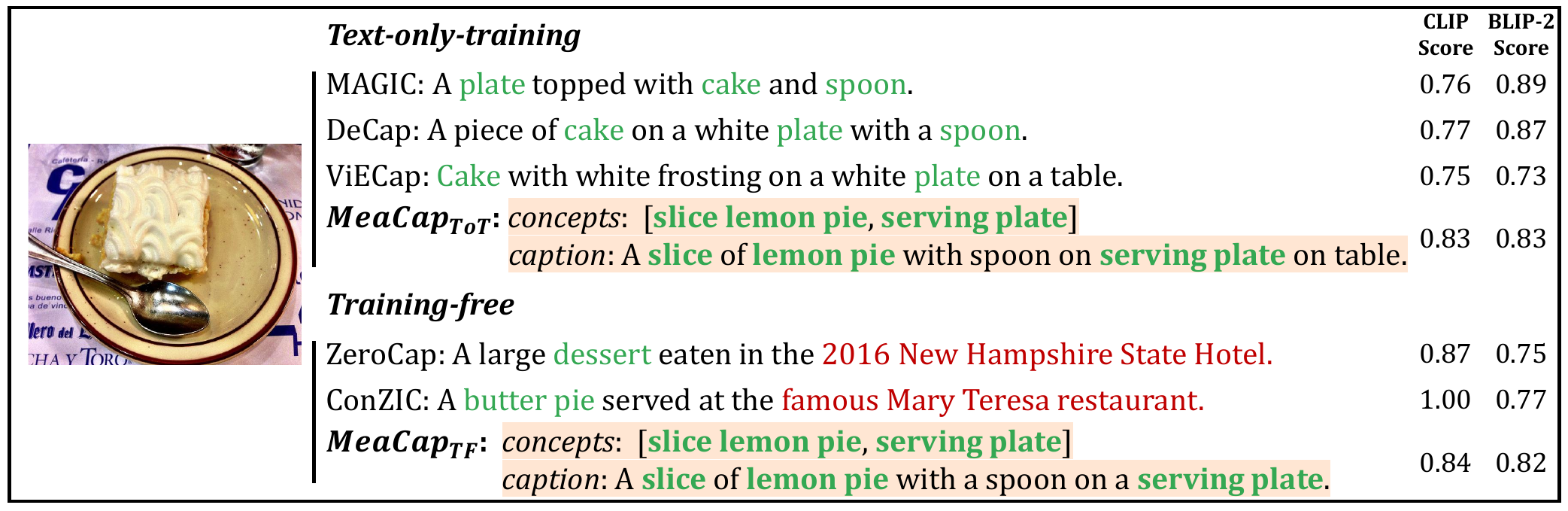}\label{fig1a}} 
  \quad
  \subfloat[Image contains world knowledge.]{\includegraphics[width=1.0\linewidth]{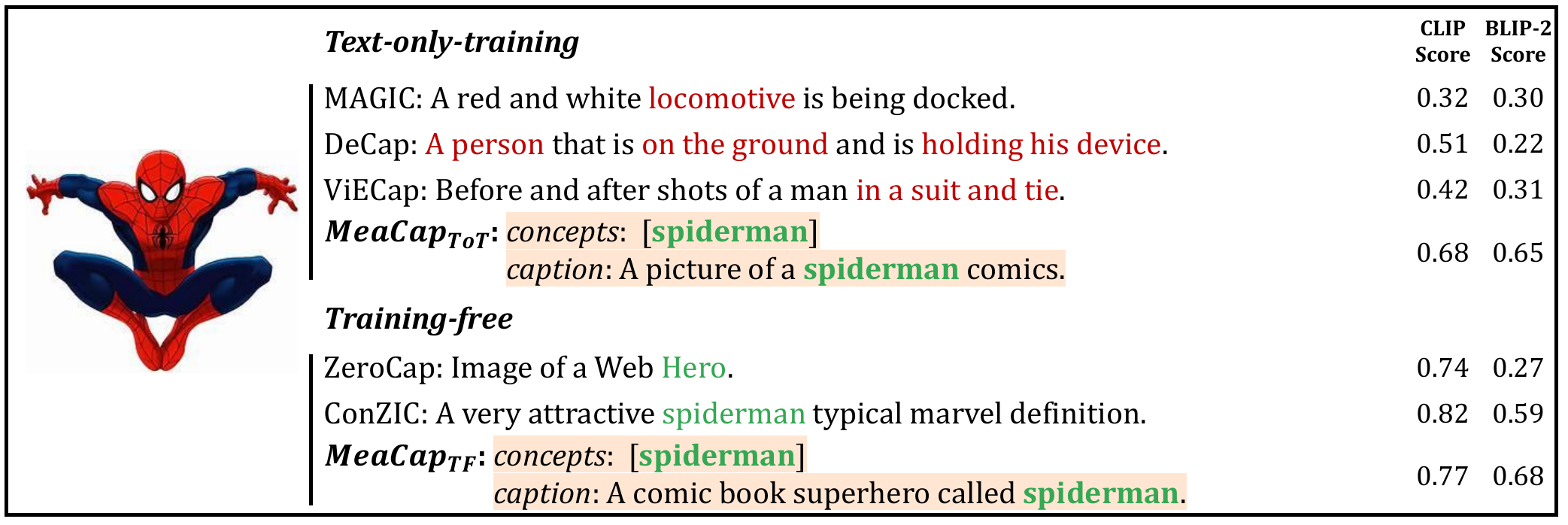}\label{fig1b}} 
    \quad
    % \subfloat[Concept-based visual-controllable captioning.]{\includegraphics[width=0.9\linewidth]{fig1c.pdf}\label{fig1c}} 
\vspace{-3mm}
  \caption{The motivation of our proposed MeaCap where the {\textcolor[RGB]{192,0,0}{red is incorrect}} and \textcolor[RGB]{52,168,83}{green is correct}. (a) Training-free methods associate the \textit{pie} with incorrect location information, which actually get high marks in CLIPscore. This might be due to the fact that CLIP is trained on web-scale noisy image-text data.
  (b) Existing text-only-training (ToT) methods fail to generate \textit{spiderman} as some training-free methods do, but the ToT version of our method (MeaCap$_{\mathrm{ToT}}$) can also do that. 
  % (c) are the visual-controllable captioning based on our proposed memory-based concepts. With the detector locating the objects in the image, we can generate \textit{object}-level captioning which focuses on specific objectives, and \textit{event}-level captioning which describes the two given objectives and their relationship.
  }
  \label{fig1}
  \vspace{-4mm}
\end{figure}
Training-free approaches \cite{tewel2022zerocap,conzic,zerocapvideo} realize zero-shot image-to-text generation using pre-trained models without fine-tuning.
Specifically, they employ a pre-trained vision-language model like CLIP %to measure the similarity between image and candidate text generated by pre-trained language model, such as Bert or GPT-2.
to guide a pre-trained language model (LM), such as BERT ~\cite{devlin2018bert} or GPT-2~\cite{gpt-2}, to generate sentences that match the given image.
With iterative inferences, this line of work does not require any training.
Though having achieved superior generalization ability and higher CLIPscore \cite{hessel2021clipscore}, these methods show extrinsic hallucination phenomenon, \ie, they tend to generate a story containing imaginary information that may not exist in the given image, as shown in Fig.~\ref{fig1a}.

To alleviate this issue, another line of works trains or fine-tunes the text decoder based on high-quality text data without corresponding images, termed as text-only-training methods \cite{magic, tu2023zerogen, li2023decap,capdec,viecap}.
For testing images containing objects described in the training corpus, text-only-training methods generate captions objectively, achieving significant improvements w.r.t. reference-based scores such as BLEU~\cite{papineni2002bleu}, METEOR~\cite{banerjee2005meteor}, and CIDEr~\cite{vedantam2015cider}.
However, due to the limited training corpus, the knowledge contained in the pre-trained LM is gradually forgotten during training, resulting in severe performance degradation on out-of-domain data, as shown in Fig.~\ref{fig1b}.
Although training on web-scale high-quality corpus is a potential solution, which hence produces extremely high computational costs.

%The textual training on captioning corpus significantly improve the accuracy performance of image captioning (similarity with human annotations, $ie$ reference-based scores BLEU~\cite{papineni2002bleu}, METEOR~\cite{banerjee2005meteor}, CIDEr~\cite{vedantam2015cider} and $\etc$), however, it also compromises the knowledge contained within the parameters of PLMs and suffers severe performance degradation on out-of-domain(OOD) data, as shown in Fig.\ref{fig1a}.
%Text-only-training methods perform well on ordinary images, but incapacity to describe those images with broader visual concepts unless they training on web-scale high-quality captioning corpus, which will impose considerable computational cost, violating the lightweight training guideline of zero-shot methods. 
% Text-only training improves faithfulness to image content and is more similar to human annotations (achieve higher human-annotation similarity, $\eg$ BLEU~\cite{papineni2002bleu}, METEOR~\cite{banerjee2005meteor}, CIDEr~\cite{vedantam2015cider}, spice~\cite{anderson2016spice}), however, it also compromises the robustness on out-of-distribution (OOD) data.
% Fig. 1(b) shows that text-only training methods are inclined to generate general captions and wrong descriptions due to related visual concepts not existing in the training corpus, $\eg$ ``Spiderman''.(overfitting to train corpus and forget knowledge from PLMs)

To maintain good generalization ability to images in the wild and to get rid of unreasonable imagination, 
this paper proposes a novel \textbf{Me}mory-\textbf{A}ugmented zero-shot image \textbf{Cap}tioning framework, namely \textbf{MeaCap}, based on the memory-guided mechanism, which provides an alternative scheme to use captioning corpus rather than using it to train the LM.
%Bearing the aforementioned limitations in mind, and to generate captions objectively of existing methods these four-aspect concerns in mind,
%\textbf{Me}mory-\textbf{A}ugmented zero-shot image \textbf{Cap}tioning framework, namely \textbf{MeaCap}.
Specifically, from an external textual memory, we develop a retrieve-then-filter module to find key concepts that are highly related to the given image.
Introducing our proposed memory-augmented visual-related fusion score to a keywords-to-sentence LM, CBART~\cite{cbart}, MeaCap can generate concept-centered captions that keep high consistency with the images. 
This new visual-related score not only considers image-text cross-modal similarity as most zero-shot IC methods~\cite{tewel2022zerocap,zerocapvideo,magic,conzic,tu2023zerogen} do by CLIP but also considers text-text in-modal similarity by evaluating the similarity between captions and retrieved image-related memory.
Our proposed MeaCap can be either training-free named {\bf{MeaCap$_{\mathrm{TF}}$}} or text-only-training named {\bf{MeaCap$_{\mathrm{ToT}}$}} by fine-tuning CBART.

%We give an alternative scheme to use textual captioning corpus instead of using the corpus to train or fine-tune the LM.
%We freeze the model parameters following training-free setting and leverage the textual captioning corpus as the external memory.
%At the inference stage, we only need to retrieve the related memory from the memory bank.
%Compared with large computation costs, the computations introduce by retrieval is negligible.
%It allows us to use the web-scale corpus to enhance the performance of zero-shot image captioning at the inference stage.
%Meanwhile, our model can seamlessly adapt to a new domain by extending new textual data to the external memory, while text-only training methods need further fine-tuning.
%Specifically, We build a retrieve-then-filter module to retrieve the most relevant texts from the external memory and then filter out the useful visual concepts. 
% Meanwhile, the filtered visual concepts can also reduce the burden on the next visual guidance
%Combined with a keywords-to-sentence LM, our model can generate concept-centered descriptions that help to keep high consistency with image content.
%Compared with previous training-free methods which only use CLIPScore as visual guidance, we further introduce a memory-augmented fusion score to further integrate visual-related linguistic guidance into the generation step of LM.  
%Moreover, we can accomplish visual-controllable captioning by combining an extra detector and our memory-based visual concepts, as shown in Fig.~\ref{fig1b}. 

Our contributions are summarized as follows:
\begin{itemize}

\item We employ the text-only captioning corpus as the external memory to enhance training-free zero-shot IC. To this end, We introduce a retrieve-then-filter module to extract key concepts from the memory and perform concept-centered generations by CBART to alleviate the hallucination issue of previous training-free methods.

%and generate concept-centered descriptions to alleviate the hallucination issue of previous training-free methods and improve the consistency of generations with image content. To explore the potential of MeaCap, we have also introduced the memory-augmented design into text-only training setting.

\item Based on the retrieved textual memory, we develop a memory-augmented visual-related fusion score into CBART, improving the correlation between image and generated captions while reserving the world-knowledge.

%visual-related linguistic information into the generation step, making the PLMs more adaptable to the captioning task and further improving the accuracy property of generated captions.

\item Extensive experiments under zero-shot, in-domain, and cross-domain scenarios demonstrate our proposed memory-augmented design can significantly improve the consistency with image content in both the training-free and text-only-training settings.

% \item By incorporating an extra detector with our extracted visual concepts, our method can achieve multi-level visual-control captioning, $\ie$ object-level captioning which focuses on specific objects, and event-level captioning which pays more attention to the relationships between two objects. 
\end{itemize}

\begin{figure*}[!tb]
	\centering 
	\includegraphics[width=0.95\textwidth]{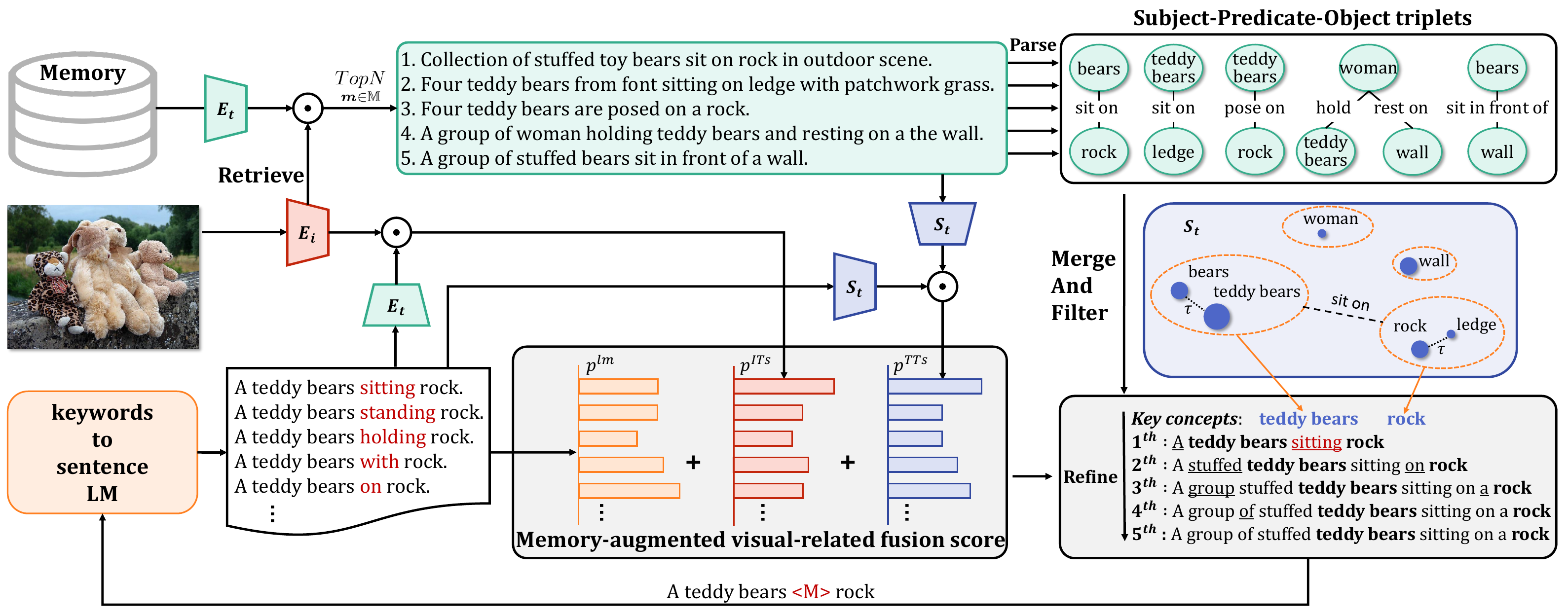}
        \vspace{-2mm}
	\caption{Overview of our proposed {\bf{MeaCap}}. The overall data flow is clockwise. \textit{i)} Given an image, we first \textit{retrieve} Top-$N$ relevant descriptions from the memory, which is transformed to the subject-predicate-object triplets; we merge and filter nodes to get the key concepts Sec.~\ref{retrieval-then-filter}. 
    \textit{ii)} With the memory-augmented visual-related fusion score (Sec.~\ref{fusion score}), starting from key concepts, the keywords-to-sentence LM can complete the image description by iterative refining (Sec.~\ref{lm}).
 $E_i, E_t, S_t$ are CLIP visual encoder, CLIP text encoder, and Sentence-BERT text encoder, respectively. \small{$\bigodot$} denotes the cosine similarity.
 The $p^{lm}, p^{ITs}, p^{TTs}$ are fluent score in Eq.~\eqref{eq:lm}, image-caption cross-modal similarity Eq.~\eqref{eq:visual}, and memory-caption in-modal similarity Eq.~\eqref{eq:memory}, respectively. 
	}\label{fig:model}
\end{figure*}
% zero-shot image captioning.
% benefit from VLM and LLM

% Modality gap: Training-free: ZeroCap, ConZIC
% problems: CLIP bias and language bias
% Fidelity: hallucinations, storytelling, LLM not design for visual captioning. 
% Text-only: magic, zerogen and CLIP-inverse(specialized language model):
% Robustness: text only training, in domain, potential limitations of LLM, world knowledge. 

% to address the issues
% external-memory-mechanism under training-free alleviates the CLIP bias and boosts the alignments between the image and the generated captions.
% 1. memory-and-filter visual concepts
% 2. memory similarity
% 3. visual controls object-level and event-level and image-level

% contribution:
% 1) introduce memory-based visual concepts, explicit injecting on late fusion and implicit injecting on Early fusion.
% 2) experiments under training-free , in domain, cross domain
% 3) visual control based on memeory-based concepts

\section{Related work}

\subsection{Supervised image captioning}
Supervised IC typically uses well-aligned image-text pairs and trains an encoder-decoder model.
For example, some early attempts~\cite{vinyals2015show, donahue2015long,xu2015show,gu2017empirical} construct 
CNN-based encoder to extract visual features and RNN/LSTM-based decoder to generate output sentence.
For better visual understanding, some methods~\cite{anderson2018bottom, cornia2020meshed, huang2019attention, huang2019adaptively, qin2019look, wang2020show, kuo2022beyond} employ an object detector to extract attentive image regions.
To encourage more interactions between two modalities, attention mechanism~\cite{cornia2020meshed,huang2019attention, nguyen2022grit, pan2020x, schwartz2019simple, schwartz2017high} and graph neural
network~\cite{yang2019auto,yao2018exploring} have been widely adopted.
%Although achieving appealing performance for in-domain case, their generaliability is not well, such as change of image style [] and novel object [].

\subsection{Zero-shot image captioning}
In recent years, zero-shot IC has gained more and more attention. 
Compared with supervised IC, zero-shot IC targets at generating image captions under two cases: \textit{i)} without any data for training named \textit{training-free} zero-shot IC; \textit{ii)} just using text from the captioning dataset to train the LM named \textit{text-only-training} zero-shot IC. 

{\bf{Training-free methods}}.
This type of methods realizes the zero-shot IC with the help of the pre-trained vision-language model, CLIP~\cite{clip}, to guide the generation of a pre-trained LM.
Specifically, ZeroCap~\cite{tewel2022zerocap} and its extension~\cite{zerocapvideo} for video captioning are proposed based on the gradient-search iteration.
To make the zero-shot IC controllable, ConZIC~\cite{conzic} is proposed by combining Gibbs-sampling with a non-autoregressive LM, improving the diversity and inference speed of IC.
Although they achieve superior generalization ability with higher CLIPscore~\cite{hessel2021clipscore}, they may generate some descriptions that do not appear in the image, called hallucination as shown in Fig.~\ref{fig1a}.

{\bf{Text-only-training methods}}.
Another line of zero-shot IC methods train or fine-tune the text decoder just using the corpus from the captioning dataset.
Concretely, after fine-tuning the off-the-shelf simCTG~\cite{simCTG} directly on the specific corpus, MAGIC~\cite{magic} and ZEROGEN~\cite{tu2023zerogen} are proposed by introducing a CLIP-induced score to regularizes the generated process of simCTG, making the caption semantically related to a given image.
By regarding the original sentence or sentence embedding as the prompt to train a LM, DeCap~\cite{li2023decap}, CapDec~\cite{capdec} and ViECap~\cite{viecap} are developed by mapping the visual feature to the text feature, which is then fed into the this LM for caption generation.

%Some works train multimodal models on web-collected noisy paired image-text data~\cite{wang2021simvlm, alayrac2022flamingo}. This solution requires considerable computation resources.
%Another solution achieves zeros-shot image captioning by combining existing VLMs and PLMs.
%Depending on the training conditions, those methods can be divided into two categories: 1) training-free methods (ZeroCap~\cite{tewel2022zerocap}, Tewel $\etal$~\cite{zerocapvideo} and ConZIC~\cite{conzic}) employ freezed VLMs to guide the PLMs towards desired visual directions via CLIP-based score, and 2) text-only-data training methods (MAGIC~\cite{magic}, ZEROGEN~\cite{tu2023zerogen}, DeCap~\cite{li2023decap}, CapDec~\cite{capdec} and ViECap~\cite{viecap}) fine-tune the PLMs on captioning corpus or train a language decoder from scratch.

Our proposed MeaCap can perform both training-free and text-only-training zero-shot IC.
For the training-free setting, because we introduce the memory mechanism for key concept identification (Sec.~\ref{retrieval-then-filter}) and guidance for LM generation (Sec.~\ref{fusion score}), our method can generate more accurate captions with less hallucination.
For text-only-training setting, the use of the corpus as the external memory in our method can alleviate the problem of existing methods that forget the world-knowledge learned by pre-trained LM due to the corpus-specific fine-tuning.

%We follow the training-free paradigm but integrate external memorys into enhanced the generations, which significantly alleviates the hallucination issues of previous training-free methods.
%We also released the text-only-data training version which fine-tuned the PLMs on the captioning corpus following previous arts, further improving the ability of keeping consistent with image content.

%To alleviate this issue, some other methods use unpaired text data to fine-tune the PLMs~\cite{magic, tu2023zerogen} or train an extra language decoder (LM)~\cite{li2023decap,capdec,viecap} to adapt the word space of LM closer to visual-related concepts. 

%use text-only data (captions without corresponding images) to fine-tune a pre-trained language model or train a text decoder from scratch.

\subsection{External memory in image captioning}
\label{sec:related memory}
It has been proven that introducing external memory is useful for various visual and language tasks, like natural language process~\cite{khandelwal2019generalization, khandelwal2020nearest, meng2021gnn, guu2020retrieval,borgeaud2022improving}, visual recognition~\cite{long2022retrieval,liu2019large}, image synthesis~\cite{duan2023few,blattmann2022retrieval}, open-domain question-answering~\cite{karpukhin2020dense,lee2019latent}, and IC included~\cite{li2023decap,ramos2023smallcap}.
For instance, SmallCap~\cite{ramos2023smallcap} is a supervised IC method that utilizes CLIP to retrieve a few relevant captions and then takes these captions as the prompt for the LM, demonstrating the memory can help LM to generate accurate captions with fewer training parameters.
In zero-shot captioning, DeCap~\cite{li2023decap} trains a LM to invert the CLIP text embeddings to the corresponding sentence. 
It projects the CLIP visual embeddings into a weighted sum of the textual memory embeddings and takes the final textual embedding as a soft prompt to the LM to guide caption generation.
% For example, for text-only zero-shot IC, DeCap~\cite{li2023decap} regards the corpus in the training set as a memory, 

%Combining deep networks with external memory is widely used in various tasks, $\eg$ natural language process (NLP)~\cite{khandelwal2019generalization, khandelwal2020nearest, meng2021gnn, guu2020retrieval,borgeaud2022improving}, visual recognition~\cite{long2022retrieval,liu2019large}, image synthesis ~\cite{duan2023few,blattmann2022retrieval}, open-domain QA~\cite{karpukhin2020dense,lee2019latent}, \textit{etc.}.
%For example, RETRO~\cite{borgeaud2022improving} proposed a memory-based transformer for language modeling that achieves comparable performance but significantly reduces the parameters and compute resources. 
%RAC~\cite{long2022retrieval} proposed an explicit retrieval module to augment long-tailed image classification. 
%VSM~\cite{duan2023few} design a variational structured memory module to boost the few-shot generation capability.
%DPR~\cite{karpukhin2020dense} implements retrieval using dense representations and establishes new state-of-the-art on multiple open-domain QA.
% In zero-shot captioning, DeCap~\cite{li2023decap} trains a language model on a textual corpus and uses the training data itself as the memory. 
% At the inference stage, DeCap projects the visual embedding to a weighted sum of memory embeddings and then takes the memory embedding as a prompt for LM to generate image descriptions.

Compared with DeCap and SmallCap, which leverage the whole memory sentence as a prompt to guide generation, we propose a training-free filter that removes the noisy information to get the key concepts from retrieved textual memory. 
Unlike the memory in DeCap is only designed for text-only-training methods, our explicit memory design can be applied to both training-free and text-only-training scenarios and shows superior capability to generate more accurate captions.

\section{MeaCap}
For better zero-shot IC with less hallucination and reserving more world-knowledge, as shown in Fig.~\ref{fig:model}, we propose a novel framework called MeaCap.
\textit{i)} To solve the problem of existing training-free methods~\cite{zerocapvideo,tewel2022zerocap,conzic} that may bring hallucination in the captions, MeaCap identifies some key concepts from the retrieved textual memory which is highly related to the image, and performs concept-centered captioning (Sec.~\ref{retrieval-then-filter}).
\textit{ii)} We develop a memory-augmented visual-related fusion score (Sec.~\ref{fusion score}), considering both image-text cross-modal similarity and text-text in-modal similarity (between textual memory and captions), which is introduced to the keywords-to-sentence LM, CBART~\cite{cbart} (Sec.~\ref{lm}), improving the image-caption correlations.

%\rr{For better zero-shot IC with less hallucination and reserving more world-knowledge, as shown in Fig.~\ref{fig:model}, we propose a novel framework called MeaCap by \textit{i)} finding key textual concepts highly related to the image from an augmented memory for removing hallucination (Sec.~\ref{retrieval-then-filter}); and \textit{ii)} updating the keywords-to-sentence LM, CBART [], to generate the complete sentence (Sec.~\ref{lm}) with world-knowledge, which is guided by our proposed visual-related fusion score (Sec.~\ref{fusion score}).}

%Our framework is shown in Fig.~\ref{fig:model}. 
%We introduce the external memory and our proposed retrieve-then-filter module in Section.~\ref{retrieval-then-filter}.
%We first utilize the image as a query to retrieve Top-$n_M$ high-similarity descriptions from the external memory by CLIP image-text similarity score. 
%Then we filter the accurate visual concepts out from the retrieved captions and remove the noisy and unrelated concepts. 
%In Section.~\ref{lm}, we introduce the keywords-to-sentence LM CBART, this pre-trained language model allows us to generate complete sentences based on visual concepts.
%In Section.~\ref{fusion score}, we propose a memory-augmented fusion score to enforce LM better generating accurate and fluent descriptions.

\subsection{Retrieve-then-filter to get key concepts}
\label{retrieval-then-filter}
The existing text-only-training zero-shot IC methods ~\cite{magic,tu2023zerogen,li2023decap,capdec,viecap} usually train or fine-tune a LM on the texts from the captioning dataset, which brings more suitable descriptions with less hallucination.
However, such methods make the generated captions overfit to a specific corpus, lacking the out-of-distribution generalization.
%\rr{However, such method will loss some important world knowledge obtained by pre-trained large VLM, which is employed in training-free zero-shot IC methods [].}
Motivated by this phenomenon, instead of training or fine-tuning a LM on the texts, we just build an augmented textual memory to get the key concepts, which can then guide the zero-shot IC.

\textbf{Build augmented memory.}
For this end, we firstly construct a large textual memory $\mathbb{M}$ which contains various visual-related sentences with abundant visual concepts.
This memory is significant for removing the hallucination for the training-free case and can alleviate the knowledge-forgotten for the text-only-training case.

\textbf{Retrieve image-related descriptions.}
Having obtained the memory, given an image $\Imat$, we use CLIP for the evaluation of image-text similarity to retrieve Top-$N$  image-related descriptions from the memory as $\{ m_n \}_{n=1}^{N_d}$: 
\begin{equation}
\label{eq: memory}
        \{m_n\}_{n=1}^{N_d} = {\mathop{TopN}_{\mv \in \mathbb{M}}} [\mathrm{cos}(E_i(\Imat), E_t(\mv))],
\end{equation}
where $E_i(\cdot)$ and $E_t(\cdot)$ denote the image and text encoder in the CLIP, respectively;
$\mathrm{cos}(\cdot, \cdot)$ is the cosine similarity.

\textbf{Subject-Predicate-Object triplets.}
To further reduce the impact of some less-information words in the image-related descriptions $\{ m_n \}_{n=1}^{N_d}$, such as article and preposition, we use an off-the-shelf textual parser, TextGraphParser~\cite{li2023factual}, to transform each description $m_n$ to a text-graph $g_n$ including multiple subject-predicate-object triplets, where subjects and objects are nodes while predicates are the relation.
These nodes are regarded as candidate concepts which will be filtered and merged to form a set of the key concepts.
The relations will decide the order between two concepts.
We define $\{ v_n \}_{n=1}^{N_c}$ as the set of all nodes from all $N_d$ text graphs $\{g_n\}_{n=1}^{N_d}$.

\textbf{Merge and filter to obtain the key concepts.}
As shown in Fig.~\ref{fig:model}, some nodes denoting concepts may represent the same object in the image  (\eg, ''bear" and ''teddy bear"), while some ones may be irrelevant to the image (\eg, ''woman"), which should be merged and filtered before getting the key concepts.

    \textit{i) Merge}.
    With the help of text encoder from Sentence-BERT~\cite{reimers-2019-sentence-bert}, $S_t(\cdot)$, we can obtain the concept embedding set $\{ \fv_n^c \}_{n=1}^{N_c}$ as $\fv_n^c = S_t(v_n)$.
    Then we evaluate the similarity between any two concept embeddings as
    \vspace{-1mm}
    \begin{equation}
        d_{ij} = \cos{(\fv_i^c, \fv_j^c)}; i,j=1,\cdots,N_c.
    \end{equation}
    Then, we set a hyper-parameter $\tau$ as the threshold, where $d_{ij} > \tau$ denotes that $i$-th concept and $j$-th concept belong to the same cluster. 
    After this, totally we have $N_v$ concept clusters as $\left\{ c_n = \{ v_i\}_{i=1}^{N_{c_n}} \right\}_{n=1}^{N_v}$, where $N_{c_n}$ denotes the number of nodes in $n$-th concept cluster $c_n$.

    \textit{2) Filter}.
    In this step, we need to decide whether the $n$-th concept cluster $c_n$ is removed or reserved.
    For this end, a reasonable assumption is that the word irrelevant to the image has a lower appearance in the retrieved descriptions  $\{ m_n \}_{n=1}^{N_d}$ in Eq.~\eqref{eq: memory}.  
    Therefore, we calculate the concept-cluster frequency $CF(c_n)$ by gradually seeing whether $v_i$ from $c_n$ appearing in $m_k$ as
    \vspace{-2mm}
    \begin{align}
        CF(c_n) &= \frac{\sum_{i=1}^{N_{c_n}} \sum_{k=1}^{N_d} \delta(v_i \in m_k)}{N_d} \\
        \delta({v_i \in m_k}) &= \left \{ \begin{aligned} 1\quad  v_i \in m_k \\ 0 \quad v_i \notin m_k \end{aligned} \right.\notag.
    \end{align}
    where $CF(c_n)$  indicates the frequency of the $n$-th cluster appearing in the retrieved descriptions $\{ m_n\}_{n=1}^{N_d}$.
    Empirically, if $CF(c_n)>0.5$, we reserve this cluster $c_n$ and otherwise delete it.
    Finally, we filter out $n_v$ key concept clusters from original $N_v$ ones, which are highly related to images.
    
    \textit{3) Find key concepts}.
    Having obtained $n_v$ key concept clusters $\{ c_n \}_{n=1}^{n_v}$ where each cluster may contain multiple similar concepts, we need to identify one concept to represent this cluster.
    For this target, we use CLIP to select one concept from one cluster by finding the maximum image-concept similarity as
    \begin{equation}
        c_n^{key} = \max_{v_j \in c_n} \left[ \mathrm{cos}(E_i(\Imat), E_t(v_j)) \right]; n=1,\cdots,n_v,
    \end{equation}
    where $c_n^{key}$ is the selected concept for the cluster $c_n$.    

After these three steps, we have the set of key concepts as $\{ c_n^{key}\}_{n=1}^{n_v}$ that is highly visual-related.
Before using these concepts to generate captions by the following keywords-to-sentence LM, we need to decide their orders, which is realized by the relations in subject-predicate-object triplets.

\subsection{Keywords-to-sentence LM}
\label{lm} 
To generate a fluent visual-related caption starting from key concepts $\{ c_n^{key}\}_{n=1}^{n_v}$, we employ a pre-trained lexically constrained language model, CBART~\cite{cbart}.
Specifically, CBART is developed to generate a sentence $S=(x_1,...,x_n)$ given the ordered $K$ keywords $\{ c_{i}\}_{i=1}^{K}$ by maximizing the conditional probability.
\begin{equation}
    S = \arg \max_{S} P(x_1,...,x_n|\{ c_{i}\}_{i=1}^{T} ),
\end{equation}
where $x_1,...,x_n$ are words.
To this end, CBART has an action encoder and a language decoder for iteratively refining the sentence starting from keywords.
At $t$-th iteration, the encoder is responsible for predicting which word-level action (\textit{copy}, \textit{replacement}, and \textit{insertion}) should be taken.
In other words, the encoder takes an incomplete sentence $S_t$ having $n'$ words as input and outputs the corresponding action sequence $L_t=\{ l_{t,1}, \cdots, l_{t,n'}\}$, where $l_{t,i}$ denotes the action of $i$-th word at $t$-th iteration.

\textit{i) Copy}.
Copy means current word remains unchanged.

\textit{ii) Replacement}.
Replacement suggests the current word should be replaced.
Specifically, CBART uses a mask token $\mathrm{<M>}$ to replace current word and sample a new word based on the conditional probability $p^{lm}(x_{\mathrm{<M>}}|x_{\mathrm{-<M>}})$, where $x_{\mathrm{-<M>}}$ denotes unmasked tokens.

\textit{iii) Insertion}.
Insertion indicates the decoder should insert a word before the current word. 
Similar to the replacement action, CBART inserts a $\mathrm{<M>}$ token before the current word and then samples a word from $p^{lm}(x_{\mathrm{<M>}}|x_{\mathrm{-<M>}})$.

Accordingly, the decoder can refine the sentence from $S_t$ to $S_{t+1}$.
Therefore, the complete encoder-decoder sentence refinement by CBART at $t$-th iteration can be formulated as
\begin{align}
    L_t = &\mathrm {LM_{Encoder}}(S_t) \\\notag
    S_{t+1} = &\mathrm {LM_{Decoder}}(S_t, L_t).
\end{align}
After a few iterations, CBART will terminate the refinement when the encoder outputs a full-copy action sequence.

According to the above introduction, existing CBART does not meet our needs because for replacement and insertion actions, the word only drawn from probability by pre-trained LM $p^{lm}(x_{\mathrm{<M>}}|x_{\mathrm{-<M>}})$, which just ensures the fluency but does not consider the visual-text relations.

\subsection{\textbf{Memory-augmented visual-related fusion score}}
\label{fusion score}
To make the captions highly-related to the given image $\Imat$, we need a visual guidance for the generation of words in the action of insertion and replacement.
Motivated by the widely-used CLIP contrastive score for evaluating the visual-text similarity, we develop a memory-augmented visual-related fusion score to adapt the original word prediction distribution of CBART to tie with the given image, considering both \textit{i) image-text cross-modal} similarity and \textit{ii) text-text in-modal} similarity.

%Up to now, we can generate complete text, but most descriptions are probably wrong because the relationships between entities and other detailed concepts come from the imagination of the language model.
%To make the generated text highly matched with the given image $I$, we need to impose visual guidance to inject image information into CBART text generation.
%Following ~\cite{magic, conzic, tu2023zerogen}, we employ a sentence-level visual-related score to adapt the original word prediction distribution to tie with the given image $I$.
Specifically, when sampling\footnote{No matter for replacement or insertion, the essence is the same, \ie, sampling a word to replace the mask.} a word $x_i$ at position $i$, CBART first predicts a conditional probability $p^{lm}$ and select top-$K_w$ candidate words $\{x_{ik}\}_{k=1}^{K_w}$ with the corresponding fluent score, as:
\vspace{-2mm}
\begin{equation}
\label{eq:lm}
p^{lm}(x_{ik}) = p^{LM}(x_{ik}|x_{-i}), k=1,\cdots,K_w 
\end{equation}
Then $K_w$ candidate sentences $\{ s_k=(x_1,...,x_{ik},...,x_n) \}_{k=1}^{K_w}$ are formed by combining candidate word $x_{ik}$ with the context $x_{-i}$.

\textit{i) image-text cross-modal similarity}.
This similarity is denoted as $p^{ITs}$, which  can be computed by taking candidate sentences $\{ s_k\}_{k=1}^{K_w}$
and the image $\Imat$ as input to calculate the CLIP cross-modality similarity as
\vspace{-2mm}
\begin{equation}
\label{eq:visual}
p^{ITs}(x_{ik}) =\mathrm{cos}(E_i(\Imat), E_t( s_k)). 
\end{equation}
% It is important to be emphasized is that the $p^{ITs}$ can not only make captions visual-related, but also introduce the \rr{world-knowledge (\eg, ''Spider-Man" in Fig.~\ref{fig1b} better than ''a man") that is learned by CLIP.}

\textit{i) text-text in-modal similarity}.
Notice that the retrieved memory $\{ m_n \}_{n=1}^{N_d}$ in Eq.~\eqref{eq: memory} is also image related.
Therefore, we introduce a memory-augmented visual-related similarity as $p^{TTs}$ to further improve the image-caption correlation by using Sentence-BERT text encoder $S_t$ to evaluate the similarity between $\{ s_k\}_{k=1}^{K_w}$ and $\{ m_n \}_{n=1}^{N_d}$ as
\vspace{-2mm}
\begin{equation}
\label{eq:memory}
    p^{TTs}(x_{ik}) = \frac{1}{N_d}\sum_{n=1}^{N_d} \mathrm{cos}(S_t(m_n), S_t(s_k)).
\end{equation}

%further introduce a memory-based score $p^{mem}$ to further impose visual-related linguistic guidance on the generated text to further improve the quality of generations, which measures the distance between candidate sentences $\{ s_k\}_{k=1}^{K_w}$ and all selected memory captions $\{m_i\}_{i=1}^{n_M}$ in the sentence embedding space, as:

Finally, after a weighted sum of Eq.~\eqref{eq:lm}, Eq.~\eqref{eq:visual} and Eq.~\eqref{eq:memory}, we have the memory-augmented visual-related fusion score as
\vspace{-2mm}
\begin{equation}
    p^{fusion} = \alphav p^{lm} + \betav p^{ITs} + \gammav p^{TTs}
\end{equation}
As a result, when sampling $i$-th word for replacement or insertion in CBART for our model, we select the candidate word with the highest fusion score as
\vspace{-2mm}
\begin{equation}
\label{eq:select}
x_i = \arg \max_{x_{ik}} p^{fusion}(x_{ik}), k=1, \cdots, K_w
\end{equation}

Up to now, our proposed MeaCap can achieve training-free zero-shot IC with less hallucination, which is named as {\bf{MeaCap$_{\mathrm{TF}}$}} in the experiments.

Moreover, Like most of text-only-training zero-shot IC models~\cite{magic, tu2023zerogen} that just use text to fine-tune the language model, we can also fine-tune the CBART firstly and then perform text-only zero-shot IC, which is named as {\bf{MeaCap$_{\mathrm{ToT}}$}} in the experiments.

%\subsection{Visual-controllable captioning}
%With the help of the detector, we can locate the $n_B$ objects with corresponding bboxes $B=\{b_i\}_{i=1}^{n_B}$ in the image $I$. For object-level captioning, we crop the sub-image $I_{b_i}$ corresponding to bbox $b_i$. Then we take the $I_{b_i}$ as an input to retrieve the most relevant visual concept $c_{b_i}$ and generate object-level captions.
%As for event-level captioning to describe the relationships between $b_i$ and $b_j$, we use extracted visual concepts $c_{b_i}$ and $c_{b_j}$ as visual concepts for this event, and then we can generate event-level captioning, as illustrated in Fig.~\ref{fig1b}. More details are shown in the Appendix.
%\label{fusion score}
\begin{table*}[htbp!]
\centering
\resizebox{\textwidth}{!}{
    \begin{tabular}{l|cc|cccccc|cccc}
        \toprule[1pt]
        \multirow{2}{*}{Methods}&\multicolumn{2}{c|}{Text Corpus} &\multicolumn{6}{c|}{MSCOCO} & \multicolumn{4}{c}{NoCap val (CIDEr)}\\
        &Training & Memory &B@4 &M & C & S &CLIP-S &BLIP2-S & In &Near &Out &Overall  \\\midrule
        ZeroCap~\cite{tewel2022zerocap} &\ding{55} & \ding{55}&2.6 &11.5 &14.6 &5.5 &\underline{0.87} &0.70 &13.3 &14.9 &19.7 &16.6\\
        Tewel $\etal$ ~\cite{zerocapvideo} &\ding{55} & \ding{55}&2.2 &12.7 &17.2 &7.3 &0.74 &0.68 &13.7 &15.8 &18.3 &16.9 \\
        ConZIC~\cite{conzic} &\ding{55} & \ding{55}&1.3 &11.2 &13.3 &5.0 &\textbf{1.00} &\underline{0.76} &15.4 &16.0 &20.3 &17.5  \\
        CLIPRe~\cite{li2023decap} &\ding{55} & CC3M &4.6 &13.3 &25.6 &9.2&0.84 &0.70 &23.3 &26.8 &36.5 &28.2 \\
        % Changpinyo $\etal$ ~\cite{cc12m} &CC3M &\ding{55} &- &- &- &- &- &- &29.2 &27.5 &37.3 &29.7 \\
        DeCap~\cite{li2023decap} &CC3M &CC3M &\underline{8.8} &16.0 &42.1 &10.9 &0.76 &- &34.8 &37.7 &\underline{49.9} &39.7   \\\midrule
        % DeCap &CC3M &COCO &- &- &- &- &- &- &- &-&- &-    \\\midrule
        \textbf{MeaCap$_{\mathrm{TF}}$} &\ding{55} &CC3M &7.1 &\underline{16.6} &\underline{42.5} &\underline{11.8} &0.84 &\textbf{0.81} &\underline{35.3} &\underline{39.0} &45.1 &\underline{40.2}  \\
      % Ours &\ding{55} &COCO &- &- &- &- &- &- &- &-&- &-  \\
          \textbf{MeaCap$_{\mathrm{ToT}}$} &CC3M &CC3M &\textbf{9.0} &\textbf{17.8} &\textbf{48.3} &\textbf{12.7} &0.79 &0.75 &\textbf{38.5} &\textbf{43.6} &\textbf{50.0} &\textbf{45.1} \\
        % Ours &CC3M &COCO &- &- &- &- &- &- &- &- &- &- \\
        \bottomrule[1pt]
    \end{tabular}}
% \end{center}
\vspace{-3mm}
\caption{Zero-shot captioning results on MSCOCO Karpathy-test split and NoCaps validations set. In, Near, and Out denote in-domain, near domain, and out-of-domain. MeaCap$_{\mathrm{TF}}$ is the training-free version and MeaCap$_{\mathrm{ToT}}$ is text-only training version.}
\vspace{-5mm}
\label{Table:zero-shot}
\end{table*}

\section{Experiments}
To demonstrate that {\bf{MeaCap}} can efficiently achieve impressive performance in different zero-shot settings, we follow the previous works~\cite{li2023decap, viecap} to conduct comprehensive experiments on \textit{Task One}: zero-shot IC in Sec.~\ref{exp:zero-shot}, and \textit{Task Two}: unpaired IC in Sec.~\ref{exp:task two}.
For each setting, we report both results of training-free version \textbf{MeaCap$_{\mathrm{TF}}$} and the text-only-training version \textbf{MeaCap$_{\mathrm{ToT}}$}.
In Sec.~\ref{exp:MeaCap-other-LM}, we further evaluate the validity of our proposed memory-based zero-shot IC framework with other LM.
In Sec.~\ref{exp:ab}, we conduct detailed ablation studies for MeaCap.

%We conduct extensive experiments to evaluate our proposed method under different scenarios:
%1) zero-shot image captioning 2) in-domain captioning 3) cross-domain captioning.
%For each setting, we both report the results of the training-free version \textbf{MeaCap$_{\mathrm{TF}}$} and the text-only training version \textbf{MeaCap$_{\mathrm{ToT}}$}.
%The experiments are organized as follows: In Sec.~\ref{exp:zero-shot}, we focused on pure zero-shot image captioning without any in-domain data and evaluated the effectiveness of our proposed memory-augmented design.
%In Sec.~\ref{exp:in-domain}, we assess the potential of MeaCap under the in-domain scenario, where the text corpus of the training set served as external memory for MeaCap$_{\mathrm{TF}}$ or the training data to fine-tune the LM for MeaCap$_{\mathrm{ToT}}$.
%In Sec.~\ref{exp:cross-domain}, we further address the transferability of MeaCap under the cross-domain setting, where the source dataset is used for training and serving as external memory, and the target dataset is used for evaluation. 
%In Sec.~\ref{exp:ab}, we conduct detailed ablation studies for MeaCap.
%In Sec.~\ref{exp:MeaCap-other-LM}, we addressed the validity of our proposed memory-based visual concepts with other LM instead of PLMs.

\textbf{Dataset}.
We conduct experiments on three widely used image captioning benchmarks, $\ie$ MSCOCO~\cite{lin2014microsoft}, Flickr30K~\cite{young2014image}, and NoCaps~\cite{agrawal2019nocaps}.
For MSCOCO and Flickr30K dataset, we follow previous works  ~\cite{cornia2020meshed,fang2022injecting,li2023decap,viecap} and use Karpathy split~\cite{karpathy2015deep}. 
We use the validation set of NoCaps to evaluate the transferability of IC models trained on other datasets.
Besides, for Task One, we follow previous works~\cite{li2023decap} that transfer the model from a web-scale corpus CC3M~\cite{cc3m} to MSCOCO and NoCaps.
CC3M contains three million image-description pairs collected from the web and we only use the text for building the memory or finetuning the LM.

\textbf{Implementation Details.}
There are various pre-trained modules used in MeaCap.
\textit{i) CLIP}: we use the pre-trained VIT-B/32 CLIP.
\textit{ii) Sentence-BERT}: we use the pre-trained model from HuggingFace\footnote{https://huggingface.co/sentence-transformers/all-MiniLM-L6-v2}.
\textit{iii) CBART}: we use the pre-trained model on One-Billion-Word corpus\footnote{https://www.statmt.org/lm-benchmark/}.
\textit{iv) TextGraphParser}: we use the off-the-shelf textual scene graph extractor~\cite{li2023factual}.
%For cross-modality similarity computation, we employ a frozen pre-trained VIT-B/32 CLIP.
%To encode the sentence into embedding space and calculate sentence similarity, we use a pre-trained Sentence-BERT\footnote{https://huggingface.co/sentence-transformers/all-MiniLM-L6-v2} ~\cite{reimers-2019-sentence-bert}.
%The language model we utilize is the CBART model pre-trained on One-Billion-Word corpus\footnote{https://www.statmt.org/lm-benchmark/}.
%To extract triplets from memory captions, we use an off-the-shelf textual scene graph extractor ~\cite{li2023factual}. 
For the training-free version MeaCap$_{\mathrm{TF}}$, we concat a prefix ``The image above depicts that'' at the start position of the sentence.
For the text-only-training version MeaCap$_{\mathrm{ToT}}$, we further fine-tune the CBART on the corresponding training corpus with AdamW~\cite{kingma2014adam} optimizer.
For Task One, we CC3M to serve as the memory, while
for Task Two, we use the training corpus of source dataset as the memory.
We set the concept similarity threshold $\tau =0.55$ for CC3M memory and $\tau =0.6$ for other memories.  
$N_d, K_w, \alphav, \betav, \gammav$ are set as $5, 200, 0.1, 0.4, 0.2$ among all experiments.
All experiments are conducted on a single RTX3090 GPU.
We preprocess the textual corpus into text embeddings by CLIP text encoder and store text embeddings as our memory for fast retrieval. 
For example, retrieval on CC3M costs an average of 0.05s on RTX3090 GPU or an average of 1s on CPU.

\textbf{Metrics}.
To evaluate the accuracy of the generated caption, we use the traditional supervised metrics BLEU (B@n)~\cite{papineni2002bleu}, METEOR (M)~\cite{banerjee2005meteor}, CIDEr (C)~\cite{vedantam2015cider}, and SPICE (S)~\cite{anderson2016spice} which compute the similarity between candidate sentences and human references.
As for training-free methods, we use the CLIPScore (CLIP-S)~\cite{hessel2021clipscore} to measure the image-text similarity.
Additionally, considering that CLIP-S is insensitive to the hallucination of those CLIP-based methods as shown in Fig.~\ref{fig1b}, we employ another pre-trained large model BLIP-2~\cite{blip2} to evaluate image-text similarity, $\ie$ BLIP2Score (BLIP2-S). More details about BLIP2-S are in the Supplement.
% For dense captioning, we use the mAP and METEOR(M) measures following ~\cite{johnson2016densecap,wu2022grit}.
\begin{figure*}[!tb]
	\centering 
	\includegraphics[width=1.0\textwidth]{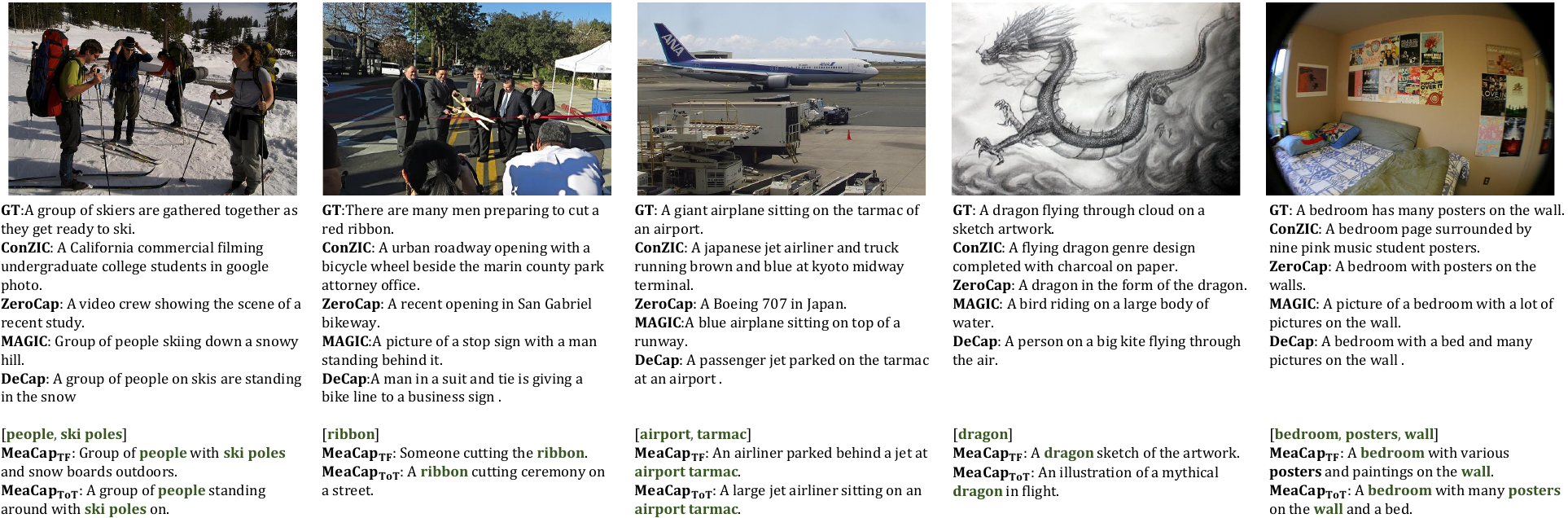}
        \vspace{-6mm}
	\caption{Examples of zero-shot IC compared with other zero-shot baselines. GT denotes the Ground Truth. ConZIC and ZeroCap are training-free, while MAGIC and DeCap are text-only-training. MeaCap displays the extracted concepts in green and generated caption.}\label{fig:cocoexamples}
 \vspace{-4mm}
\end{figure*}

\begin{figure}[!tb]
	\centering 
	\includegraphics[width=0.5\textwidth]{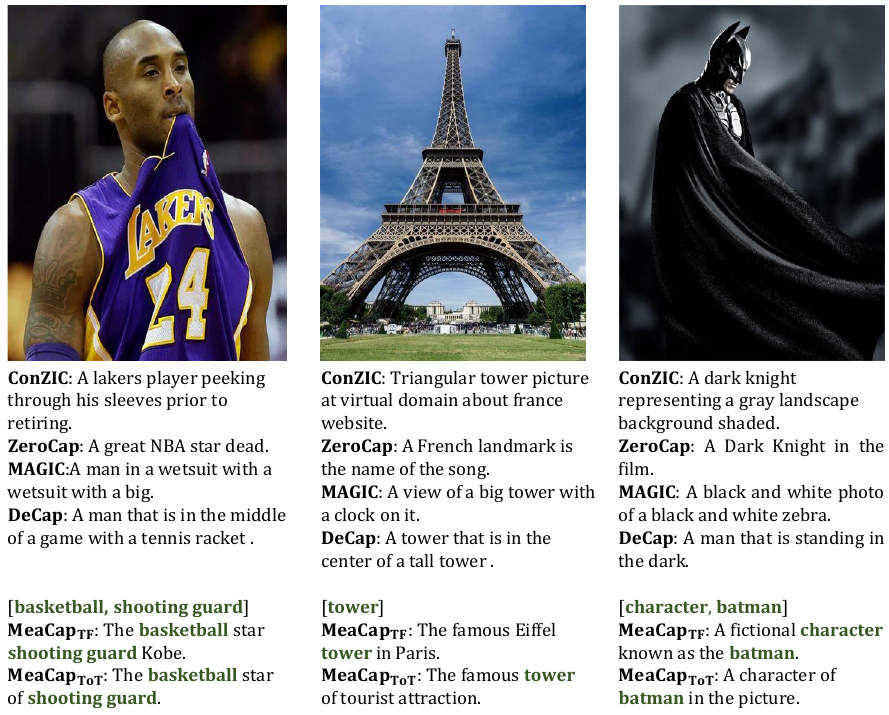}
        \vspace{-5mm}
	\caption{Examples of real-world knowledge. MeaCap$_\mathrm{ToT}$ can alleviate the world-knowledge-forgotten problem of existing text-only-training methods, such as ''batman" in the third image.}\label{fig:knowledge}
        \vspace{-4mm}
\end{figure}

\begin{table}[tp!]
\centering
\resizebox{0.5\textwidth}{!}{
    \begin{tabular}{l|cccc|cccc}
        \toprule[1pt]
        \multirow{2}{*}{Methods}&\multicolumn{4}{c|}{MSCOCO} & \multicolumn{4}{c}{Flickr30K}\\
         &B@4 &M & C & S &B@4 &M & C & S  \\\midrule
        &\multicolumn{8}{c}{\gg{Training on image-text pairs}} \\\midrule
        \gg{Bottom-Up}~\cite{anderson2018bottom} &\gg{36.2} &\gg{27.0} &\gg{113.5} &\gg{20.3} &\gg{27.3} & \gg{21.7} &\gg{56.6} &\gg{16.0}\\
        \gg{OSCAR}~\cite{li2020oscar} &\gg{36.5} &\gg{30.3} &\gg{123.7} &\gg{23.1} &\gg{-} &\gg{-} &\gg{-} &\gg{-} \\
        \gg{VinVL}~\cite{zhang2021vinvl} &\gg{40.9} &\gg{30.9} &\gg{140.6} &\gg{25.1} &\gg{-} &\gg{-} &\gg{-} &\gg{-} \\
        \gg{ClipCap}~\cite{mokady2021clipcap}&\gg{33.5} &\gg{27.5} &\gg{113.1} &\gg{21.1} &\gg{-} &\gg{-} &\gg{-} &\gg{-} \\
        \gg{SmallCap}~\cite{ramos2023smallcap} &\gg{37.0} &\gg{27.9} &\gg{119.7} &\gg{21.3}&\gg{-} &\gg{-} &\gg{-} &\gg{-} \\
        \gg{I-Tuning}~\cite{luo2023tuning} &\gg{34.8} &\gg{28.3} &\gg{116.7} &\gg{21.8} &\gg{25.2} &\gg{22.8} &\gg{61.5} &\gg{16.9} \\\midrule
        &\multicolumn{8}{c}{Text-only-training, zero-shot inference} \\\midrule
        ZeroCap$\dagger$~\cite{tewel2022zerocap}&7.0 &15.4 &49.3&9.2 &5.4 &11.8 &16.8 &6.2 \\
        MAGIC~\cite{magic}&12.9 &17.4 &49.3 &11.3 &6.4 &13.1 &20.4 &7.1 \\
        % ConZIC&- &- &- &-&- &- &- &- \\
        ZEROGEN~\cite{tu2023zerogen}&\underline{15.5} &18.7 &55.4&12.1 &\underline{13.1} &15.2 &26.4 &8.3 \\
        CLIPRe~\cite{li2023decap}&12.4 &20.4 &53.4&14.8 &9.8 &\underline{18.2} &31.7 &12.0 \\\midrule
        \textbf{MeaCap$_{\mathrm{TF}}$}  &9.1 &\underline{20.6} &\underline{56.9} &\underline{15.5} &7.2 &17.8 &\underline{36.5} &\underline{13.1}  \\
         \textbf{MeaCap$_{\mathrm{ToT}}$}  &\textbf{17.7} &\textbf{24.3} &\textbf{84.8} &\textbf{18.7} &\textbf{15.3} &\textbf{20.6} &\textbf{50.2} &\textbf{14.5}  \\
        \bottomrule[1pt]
    \end{tabular}}
% \end{center}
\vspace{-2mm}
\caption{In-domain captioning results on MSCOCO Karpathy-test split and Flickr30K Karpathy-test split. $\dagger$ means text-only re-implemented version from \cite{magic}.}
\vspace{-4mm}
\label{Table:in-domain}
\end{table}

\subsection{Zero-shot image captioning}
\label{exp:zero-shot}
In this section, we conduct zero-shot IC experiments to evaluate the ability of models to transfer from a general web-collected corpus to different downstream IC datasets.

\textbf{Baselines.}
In this study, we compare two types of baselines. \textit{i)} Training-free methods: ZeroCap~\cite{tewel2022zerocap}, Tewel $\etal$~\cite{zerocapvideo} and ConZIC~\cite{conzic}.
Those methods leverage pre-trained CLIP and freezed LM (BERT or GPT-2) to achieve zero-shot IC.
\textit{ii)} Text-only-training methods\footnote{Other text-only-training methods except DeCap do not have experimented in this setting, we compared them in Task Two (Sec.~\ref{exp:task two})}: DeCap~\cite{li2023decap}, which is also a memory-based method discussed in Sec.~\ref{sec:related memory}.
Instead of using pre-trained LM, DeCap trains a language decoder from scratch.
Besides, authors of Decap set up a baseline called CLIPRe, which generate image descriptions by retrieving the most relevant texts from memory directly.
Following Decap, for MeaCap$_\mathrm{TF}$, we just use CC3M as the memory, and for MeaCap$_\mathrm{ToT}$, we use CC3M as the memory and also tuning the CBART.
Tab.~\ref{Table:zero-shot} shows the results on MSCOCO and NoCaps, and MeaCap achieves new state-of-the-art results.
%We evaluate the MeaCap by training-free  textbf{MeaCap$_\mathrm{TF}$} and text-ony, which means no training process is involved and all modules are pre-trained.
% To further explore the potential of MeaCap, we fine-tuned the LM module on the CC3M text corpus to adapt the LM to better fit the captioning task, $\ie$ the text-only version \textbf{MeaCap$_\mathrm{TOT}$}.
\begin{table}[tp!]
\centering
\resizebox{0.5\textwidth}{!}{
    \begin{tabular}{l|cccc|cccc}
        \toprule[1pt]
        \multirow{2}{*}{Methods}&\multicolumn{4}{c|}{MSCOCO $\rightarrow$ Flickr30k} & \multicolumn{4}{c}{Flickr30k$\rightarrow$MSCOCO}\\
         &B@4 &M & C & S &B@4 &M & C & S  \\\midrule
        MAGIC~\cite{magic} &6.2 &12.2 &17.5 &5.9  &5.2 &12.5 &18.3 &5.7 \\
        CLIPRe~\cite{li2023decap} &\underline{9.8} &\underline{16.7} &30.1 &10.3  &6.0 &16.0 &26.5 &10.2 \\\midrule
        % DeCap &16.3 &17.9 &35.7 &11.1 &12.1 &18.0 &44.4 &10.9 \\
        % CapDec &17.3 &18.6 &35.7 &- &9.2 &16.3 &27.3 &-  \\
        % ViECap  &17.4 &18.0 &38.4 &11.2 &12.6 &19.3 &54.2 &12.5  \\\midrule
        \textbf{MeaCap$_{\mathrm{TF}}$}  &7.1 &16.6 &\underline{34.4} &\underline{11.4} &\underline{7.4} &\underline{16.2} &\underline{46.4} &\underline{11.2}  \\
        % \textbf{MeaCap$_{\mathrm{TOT+CC3M}}$}  &\textbf{13.5} &17.5 &40.0 &\textbf{12.1} &8.9 &16.2 &43.8 &11.5  \\
        \textbf{MeaCap$_{\mathrm{ToT}}$}  &\textbf{13.4} &\textbf{18.5} &\textbf{40.3} &\textbf{12.1} &\textbf{9.8} &\textbf{17.4} &\textbf{51.7} &\textbf{12.0}  \\
        \bottomrule[1pt]
    \end{tabular}}
    \vspace{-3mm}
% \end{center}
\caption{Cross domain captioning results on MSCOCO and Flickr30K Karpathy-test split.}
\label{Table:cross-domain}
\vspace{-5mm}
\end{table}
\textbf{Training-free Results.}
Concretely, our training-free version MeaCap$_\mathrm{TF}$ has shown superior performance on reference-based metric (B@4, M, C, S)  than all previous training-free baselines, ZeroCap, Tewel $\etal$ and ConZIC on both MSCOCO and NoCaps datasets by a large margin, demonstrating the effectiveness of our memory-augmented design. 
For reference-free metrics (CLIP-S and BLIP2-S), MeaCap$_\mathrm{TF}$ achieves better results on BLIP2-S and is inferior on CLIP-S.
As discussed in the introduction, previous training-free methods are favored by CLIP-S because of the hallucination phenomenon. 
%BLIP2-S may be a more fair in that our method and previous baselines are all CLIP-based methods.
Besides, Our MeaCap$_\mathrm{TF}$ also surpasses the retrieval-based baseline CLIPRe by a large margin, indicating that only retrieving the most relevant caption is deficient in accuracy. 
%Removing the noisy and incorrect concepts is critical for zero-shot IC.
Moreover, even compared with the text-only-training method Decap, MeaCap$_\mathrm{TF}$ shows superior or comparable performance on both MSCOCO and NoCap.

%compared with the previous memory-based text-only-training method DeCap, MeaCap$_\mathrm{TF}$ shows superior and comparable performance (+0.6, +3.3, +0.9 improvements on M, C, and S on the MSCOCO dataset). 
% Notably, DeCap uses CC3M textual corpus to train the language model instead of employing pre-trained PLMs, which gives it an edge over reference-based metrics, $\ie$ B@4, M, C, and S.
\textbf{Text-only-training Results.}
To explore the potential of our MeaCap with further text-only-training on the web-scale corpus following DeCap, we also fine-tune CBART on CC3M corpus, $\ie$ MeaCap$_\mathrm{ToT}$.
It can be observed that MeaCap$_\mathrm{ToT}$ significantly improves the performance, especially on NoCap.
Specifically, under the same training and memory condition, MeaCap$_\mathrm{ToT}$ surpasses DeCap in both the MSCOCO dataset and the NoCaps dataset, 
%(MSCOCO +1.8, +6.2, +1.8 improvements on M, C, and S and NoCaps +4.6, +5.5, +3.8 on CIDEr under in-domain, near-domain, and out-of-domain setting)
showing the superiority of our method to use the external memory.

\textbf{Qualitative results.}
Besides quantitative compare, we visualize the generated captions in Fig.~\ref{fig1}, Fig.~\ref{fig:cocoexamples}, and Fig.~\ref{fig:knowledge}.
Clearly, our proposed MeaCap can achieve better caption with more knowledge and less hallucination.

%the visualized results indicate that MeaCAP can generate accurate descriptions.
%Fig.~\ref{fig:cocoexamples} and Fig.~\ref{fig:knowledge} show some qualitative results.
%Our method achieves better captioning results compared with previous zero-shot captioning methods.

\subsection{Task Two: Unpaired image captioning}
\label{exp:task two}
\vspace{-1mm}
\subsubsection{In-domain captioning}
\vspace{-2mm}
To explore more potential of MeaCap under an in-domain setting, which means the training data, the memory, and the test set are from the same dataset, but do not use image-text pairs to build the model and memory.

\textbf{Baselines.}
In this study, we compare with other text-only-training methods ZeroCap$\dagger$~\cite{tewel2022zerocap}, MAGIC~\cite{magic}, and ZEROGEN~\cite{tu2023zerogen} and a retrieval-based approach CLIPRe.
ZeroCap is a training-free method which is extended to text-only-training version ZeroCap$\dagger$~\cite{tewel2022zerocap}.
Those methods freeze the CLIP and fine-tune the LM on corresponding training texts.  
Under the in-domain setting, we also report both the training-free version MeaCap$_\mathrm{TF}$, which only employs the training text as memory, and the text-only-training version MeaCap$_\mathrm{ToT}$ which utilizes the training text to fine-tune CBART and serve as memory as well.

\textbf{Results.}
As shown in Tab.~\ref{Table:in-domain}, MeaCap$_\mathrm{TF}$ outperforms CLIPRe and other text-only-training baselines on C and S scores. Compared with B@4 and M scores, The C and S scores pay more attention to the accuracy of entities and relationships. The superior performance on these two scores demonstrates the high quality of our proposed memory-based retrieval-then-filter method to get the key concepts.
Moreover, MeaCap$_\mathrm{ToT}$ outperforms all baselines by a large margin, indicating that our proposed method has greater potential with further in-domain training.
% It indicates that incorporating the memory-augmented design is beneficial for generating accurate and fluent descriptions.
\vspace{-4mm}
\subsubsection{Cross-domain captioning}
\label{exp:cross-domain}
\vspace{-2mm}
We evaluate the MeaCap in a cross-domain setting where the training data and testing data are from different datasets. We use the text from the source domain as the memory for MeaCap$_\mathrm{TF}$ and MeaCap$_\mathrm{ToT}$, and fine-tune the CBART for MeaCap$_\mathrm{ToT}$.

\textbf{Results.}
We compare our method with the text-only-training baseline MAGIC which fine-tune the GPT-2, and a retrieval-based baseline CLIPRe. 
Results in Tab.~\ref{Table:cross-domain} show that MAGIC with finetuning the GPT-2 on source data suffers a performance degradation on target data, even worse than the retrieval-based method CLIPRe. 
% The same with MAGIC, our method is also based on a pre-trained LM. 
Equipped with the proposed memory-augmented design, MeaCap$_\mathrm{TF}$ surpasses the CLIPRe on most metrics and MeaCap$_\mathrm{ToT}$ outperforms all baselines a lot, demonstrating the effectiveness of the proposed memory-augmented design.

\begin{figure}[!tb]
	\centering 
	\includegraphics[width=0.5\textwidth]{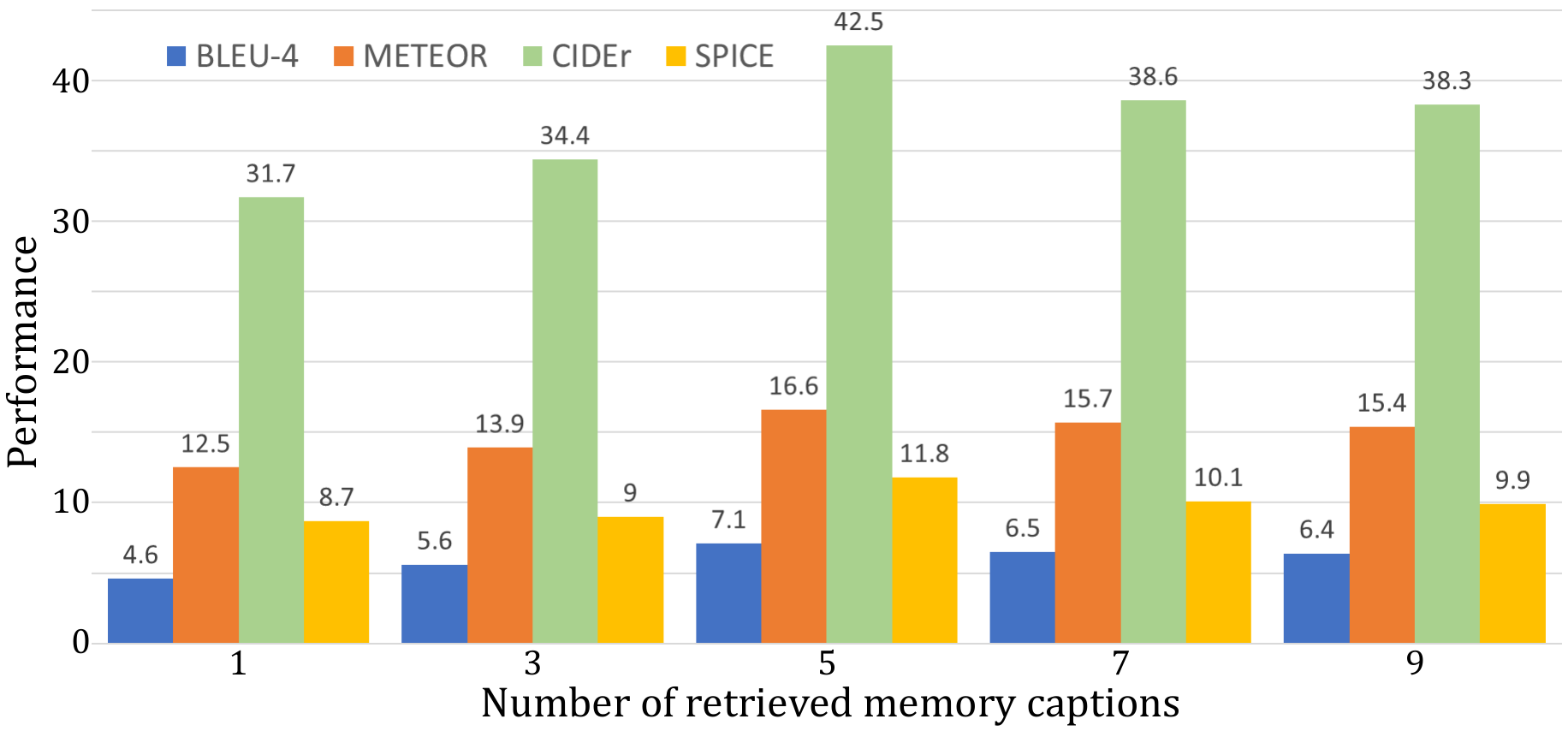}
        \vspace{-6mm}
	\caption{Effect of the number of retrieved memory captions. We reported the performance of $\mathrm{MeaCap_{TF}}$ on the MSCOCO dataset with varying the number of retrieved memory captions.}\label{fig:ablation}
 \vspace{-2mm}
\end{figure}

\begin{table}[tp!]
\centering
\resizebox{0.5\textwidth}{!}{
    \begin{tabular}{l|cccc|cccc}
        \toprule[1pt]
        \multirow{2}{*}{Methods}&\multicolumn{4}{c|}{MSCOCO} & \multicolumn{4}{c}{Flickr30K}\\
        &B@4 &M & C & S &B@4 &M & C & S  \\\midrule
        DeCap~\cite{li2023decap} &24.7 &25.0 &91.2 &\underline{18.7} &21.2 &\underline{21.8} &\underline{56.7} &\underline{15.2} \\
        CapDec~\cite{capdec} &26.4 &25.1 &91.8 &- &17.7 &20.0 &39.1 &-  \\
        ViECap~\cite{viecap}  &\textbf{27.2} &\underline{24.8} &\underline{92.9} &18.2 &\underline{21.4} &20.1 &47.9 &13.6   \\
        \textbf{MeaCap$_{\mathrm{InvLM}}$}  &\textbf{27.2} &\textbf{25.3} &\textbf{95.4} &\textbf{19.0} &\textbf{22.3} &\textbf{22.3} &\textbf{59.4} &\textbf{15.6}  \\\midrule
        &\multicolumn{4}{c|}{MSCOCO $\rightarrow$ Flickr30K} & \multicolumn{4}{c}{Flickr30K$\rightarrow$MSCOCO}\\\midrule
        DeCap~\cite{li2023decap} &16.3 &17.9 &35.7 &11.1 &12.1 &18.0 &44.4 &10.9 \\
        CapDec~\cite{capdec} &17.3 &\underline{18.6} &35.7 &- &9.2 &16.3 &27.3 &-  \\
        ViECap~\cite{viecap}  &\underline{17.4} &18.0 &\underline{38.4} &\underline{11.2} &\underline{12.6} &\underline{19.3} &\underline{54.2} &\underline{12.5}   \\
        \textbf{MeaCap$_{\mathrm{InvLM}}$}  &\textbf{18.5} &\textbf{19.5} &\textbf{43.9} 
        &\textbf{12.8} &\textbf{13.1} &\textbf{19.7} &\textbf{56.4} &\textbf{13.2}  \\
        \bottomrule[1pt]
    \end{tabular}}
% \end{center}
\vspace{-3mm}
\caption{In-domain and cross-domain captioning results with CLIP-invert language decoder. }
\label{Table:clip-invert}
\vspace{-3mm}
\end{table}
\subsection{Flexibility of MeaCap with other LM}
\label{exp:MeaCap-other-LM}
It is important to emphasize that our proposed memory mechanism for finding key concepts in Sec.~\ref{retrieval-then-filter} is a plug-and-play module to further improve most of the existing text-only-training SOTA methods~\cite{li2023decap, capdec, viecap} on Task Two.
%For the zero-shot setting of Task Two in Sec.~\ref{exp:task two}, just fine-tuning the CBART is not sufficient to obtain the SOTA performance compared with some methods~\cite{li2023decap, capdec, viecap}, which design a task-specifc LM and train it from scratch.
%This is because Task Two requires the LM to be more focused on one specific dataset than Task One.
%Thanks to the \textit{plug-and-play} property of our proposed memory mechanism for finding key concepts in Sec.~\ref{retrieval-then-filter}
For this end, we just replace the CBART (Sec.~\ref{lm}) in current MeaCap with the another LM used in these methods~\cite{li2023decap, capdec, viecap} (also do not need to use the fusion score in Sec.~\ref{fusion score}), which is described detailed below.

%To address the validity of our proposed memory-augmented design, we extend MeaCap to the LM training-from-scratch following previous methods DeCap~\cite{li2023decap}, CapDec~\cite{capdec}, ViECap~\cite{viecap}, instead of leveraging PLMs. 
\textbf{Baselines.}
DeCap~\cite{li2023decap}, CapDec~\cite{capdec} and ViECap~\cite{viecap} train a LM from scratch to invert the CLIP text encoder, denoted as InvLM in the following. 
They project the visual embeddings extracted by the CLIP visual encoder to the text embedding space of the CLIP text encoder.
Then, they use InvLM to reconstruct the text from text embeddings.
To generate descriptions based on our extracted key concepts, we first use a prompt template as ``There are [$c_1,c_2,...,c_n$] in the image'' to inject the concepts into a concept-aware sentence following ViECap, where $c_n$ are the $n$-th concepts.
After encoding the concept-aware sentence to text embeddings by CLIP text encoder, we get a concept-aware prompt.
We concat the concept-aware prompt with textual embeddings as the input of InvLM.
We call this version MeaCap as \textbf{MeaCap$_{\mathrm{InvLM}}$}

\textbf{Results.}
Tab.~\ref{Table:clip-invert} shows that MeaCap$_\mathrm{{InvLM}}$ outperforms all baselines on all metrics under in-domain and cross-domain scenarios, demonstrating the effectiveness of our proposed memory-based key concepts, and also indicating its flexibility for various LM and different zero-shot settings.
More details are shown in the Supplement.

\subsection{Ablation studies}
\label{exp:ab}
To explore the impact of each key module in MeaCap, $\ie$ the retrieve-then-filter module (\textbf{ReF}), the image-text similarity score (\textbf{ITs}), and the text-text similarity score (\textbf{TTs}), we conduct comprehensive ablation studies on the MSCOCO dataset based on the Task One of zero-shot setting.
We evaluate both the training-free version MeaCap$_\mathrm{TF}$ and the text-only training version MeaCap$_\mathrm{ToT}$ whose results are provided in Tab.~\ref{Table:ab}.
As we can see, only combined with the ReF and original LM (the first row) can surpass the only ITs results in the second row (ITs is the only visual guidance of previous training-free methods by CLIP), indicating the key concepts extracted by the ReF module are critical for zero-shot IC.
The third row shows that combining ReF with ITs yields more improvements than individual modules alone.
Finally, by incorporating the TTs, the performance is further improved, highlighting the efficacy of the memory-augmented visual-related fusion score.
We also conduct experiments to investigate the impact of the number of retrieved memory captions as shown in Fig.~\ref{fig:ablation}. 
Our model achieves best performance when retrieving five memory captions.

\begin{table}[tp!]
\centering
\resizebox{0.45\textwidth}{!}{
    \begin{tabular}{l|ccc|cccc}
        \toprule[1pt]
         \multirow{2}{*}{Methods}&\multirow{2}{*}{\textbf{ReF}} &\multirow{2}{*}{\textbf{ITs}} & \multirow{2}{*}{\textbf{TTs}}& \multicolumn{4}{c}{MSCOCO}\\
         &&& &B@4 &M & C & S  \\\midrule
        \multirow{4}{*}{MeaCap$_{\mathrm{TF}}$}&\ding{52} &\ding{55} &\ding{55} &5.0 &13.3 &31.1 &5.6  \\
        &\ding{55} &\ding{52} &\ding{55} &1.8 &9.7 &12.7 &4.8  \\
        &\ding{52} &\ding{52} &\ding{55} &5.7 &13.6 &38.6 &8.5    \\
        &\ding{52} &\ding{52} &\ding{52} &7.1 &16.6 &42.5 &11.8    \\\midrule
        \multirow{4}{*}{{MeaCap$_{\mathrm{ToT}}$}}&\ding{52} &\ding{55} &\ding{55} &7.9 &14.9 &37.1 &10.4  \\
        &\ding{55} &\ding{52} &\ding{55} &3.2 &9.9 &17.3 &5.2  \\
        &\ding{52} &\ding{52} &\ding{55} &8.1 &15.6 &44.7 &11.1    \\
        &\ding{52} &\ding{52} &\ding{52} &9.0 &17.8 &48.3 &12.7    \\
        \bottomrule[1pt]
    \end{tabular}}
% \end{center}
\vspace{-2mm}
\caption{Ablation studies on zero-shot image captioning. ReF, ITs, TTs denote the retrieve-and-filter module, ITs~\eqref{eq:visual} and TTs~\eqref{eq: memory} are image-text similarity and text-text similarity from our memory-augmented visual-related score, respectively. }
% \vspace{-3mm}
\label{Table:ab}
\end{table}

\section{Conclusion}
In this paper, we propose a novel memory-augmented zero-shot image captioning framework, namely MeaCap. We introduce a retrieve-then-filter module to extract key concepts from external textual memory. Based on the retrieved textual memory, we further develop a memory-augmented visual-related fusion score to guide the generation of captions. Combined with CBART, we can generate concept-centered descriptions to alleviate the hallucination of previous training-free methods and enhance the accuracy of text-only-training methods.
Extensive experiments on various zero-shot captioning settings show that our proposed MeaCap outperforms previous training-free and text-only-training methods.

% \clearpage
% \newpage
{
    \small
    \bibliographystyle{ieeenat_fullname}
    \bibliography{main}
}
\clearpage
\appendix
\begin{figure}[!ht]
	\centering 
	\includegraphics[width=0.45\textwidth]{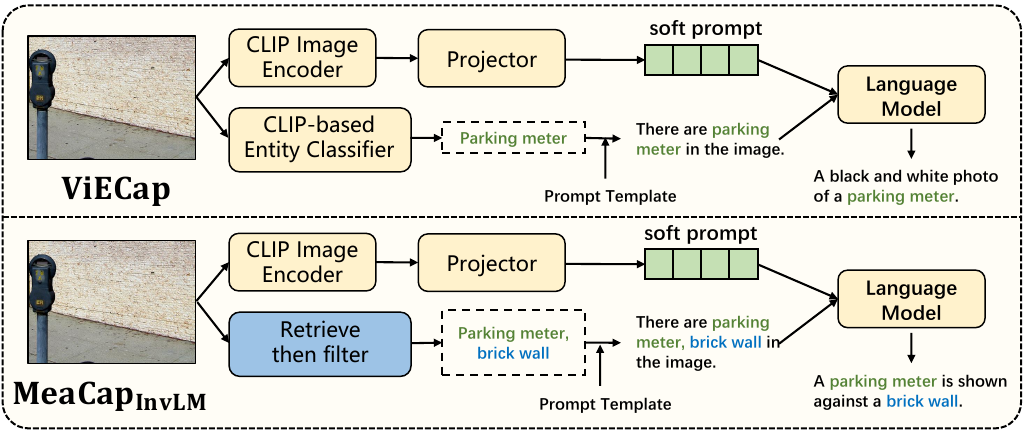}
        \vspace{-2mm}
	\caption{Difference between ViECap~\cite{viecap} and MeaCap$_\mathrm{InvLM}$.}\label{fig:viecap}
\end{figure}

\begin{figure}[!t]
	\centering 
	\includegraphics[width=0.45\textwidth]{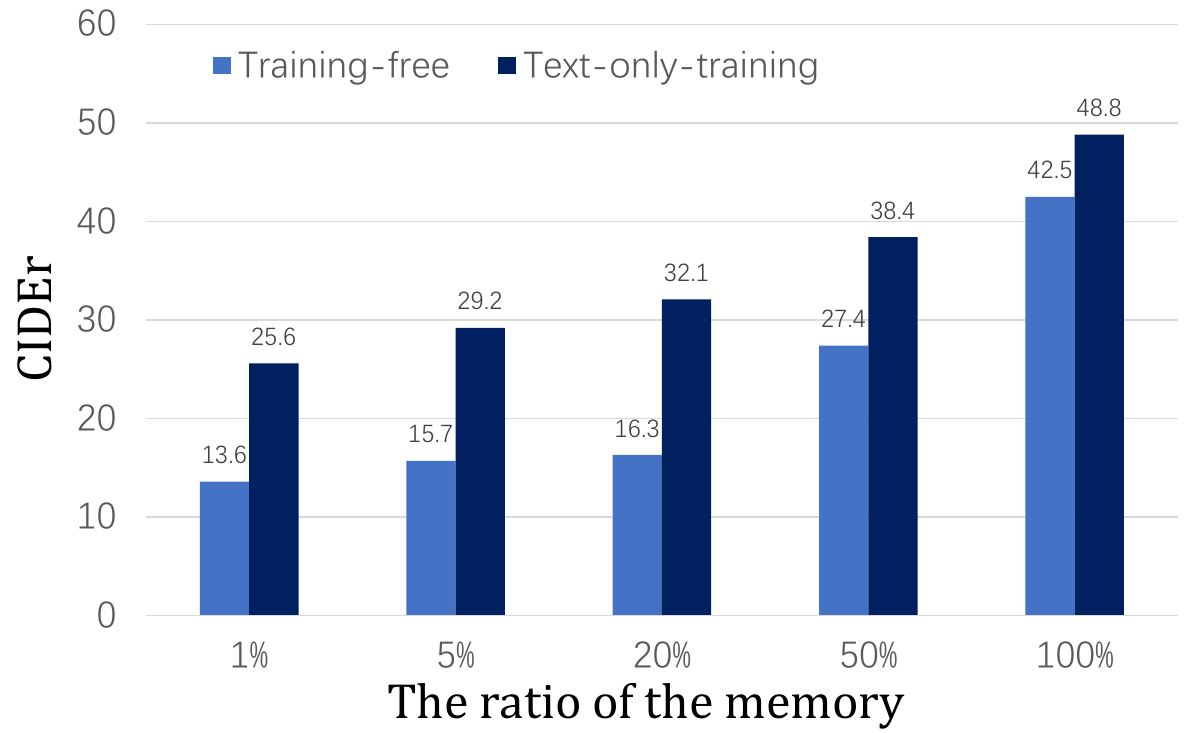}
        \vspace{-2mm}
	\caption{Ablation study on memory size.}\label{fig:memory_size}
\end{figure}
\section{More details of MeaCap with other LM}
To demonstrate the effectiveness of our proposed memory concepts, we incorporate our proposed retrieve-then-filter module with the current SOTA text-only-training method ViECap~\cite{viecap}, namely MeaCap$_{\mathrm{InvLM}}$.
ViECap is a text-only-training method that combines entity-aware hard prompts and learned visual soft prompts to generate image captions. 
The soft prompts are obtained by inputting the CLIP image embedding into the projector and the hard prompts are constructed by a CLIP-based entity classifier.
As illustrated in Fig.~\ref{fig:viecap}, MeaCap$_{\mathrm{InvLM}}$ remains the soft prompts branch and directly utilizes the pre-trained language model of ViECap.
We only replace the entity classifier with our proposed retrieve-the-filter module to obtain key concepts and leverage the prompt template to inject the key concepts into a concept-aware sentence and get the hard prompts.
Compared with the original ViECap which can only retrieve a single entity from pre-defined entity vocabulary, our proposed retrieve-then-filter module can extract more key visual concepts from the image, demonstrating that our proposed memory design can help to generate highly consistent descriptions with image content.
Quantitative results under in-domain and cross-domain settings of MeaCap$_{\mathrm{InvLM}}$ are shown in Tab. 4 in the main paper.

\section{The size of memory}
To explore the impact of the external memory size, we randomly sample different ratio text embeddings of CC3M as our external memory.
The results are shown in Fig.~\ref{fig:memory_size}.
Overall, the training-free MeaCap$_\mathrm{TF}$ and text-only-training MeaCap$_\mathrm{ToT}$ both benefit from a large memory, and the performance increases as the memory size grows.
It is due to that large memory usually covers more visual concepts and thus the retrieved captions are better aligned with the same image.

\section{BLIP2-S}
BLIP-2~\cite{blip2} is a large vision-language model that employs a frozen CLIP~\cite{clip} image encoder and large language models~\cite{flan,zhang2022opt} to bootstrapping language-image pre-training on large image-text corpus.
We use the pre-trained BLIP-2 and leverage the image encoder $B_v$ and the text encoder $B_t$ to compute cross-modality cosine similarity.
Following CLIP-S~\cite{hessel2021clipscore}, BLIP2-S is the product of a weight $w=2$ and the image-text similarity, as:
\begin{equation}
    \mathrm{BLIP2-S}(\Imat, \Tmat)=w\mathrm{Cos}(B_v(\Imat), B_t(\Tmat))
\end{equation}
where $\Imat$, and $\Tmat$ are the image and the text. $\mathrm{Cos}$ denotes the cosine similarity.

\section{More samples and qualitatives.}
\begin{figure*}[!tb]
	\centering 
	\includegraphics[width=0.95\textwidth]{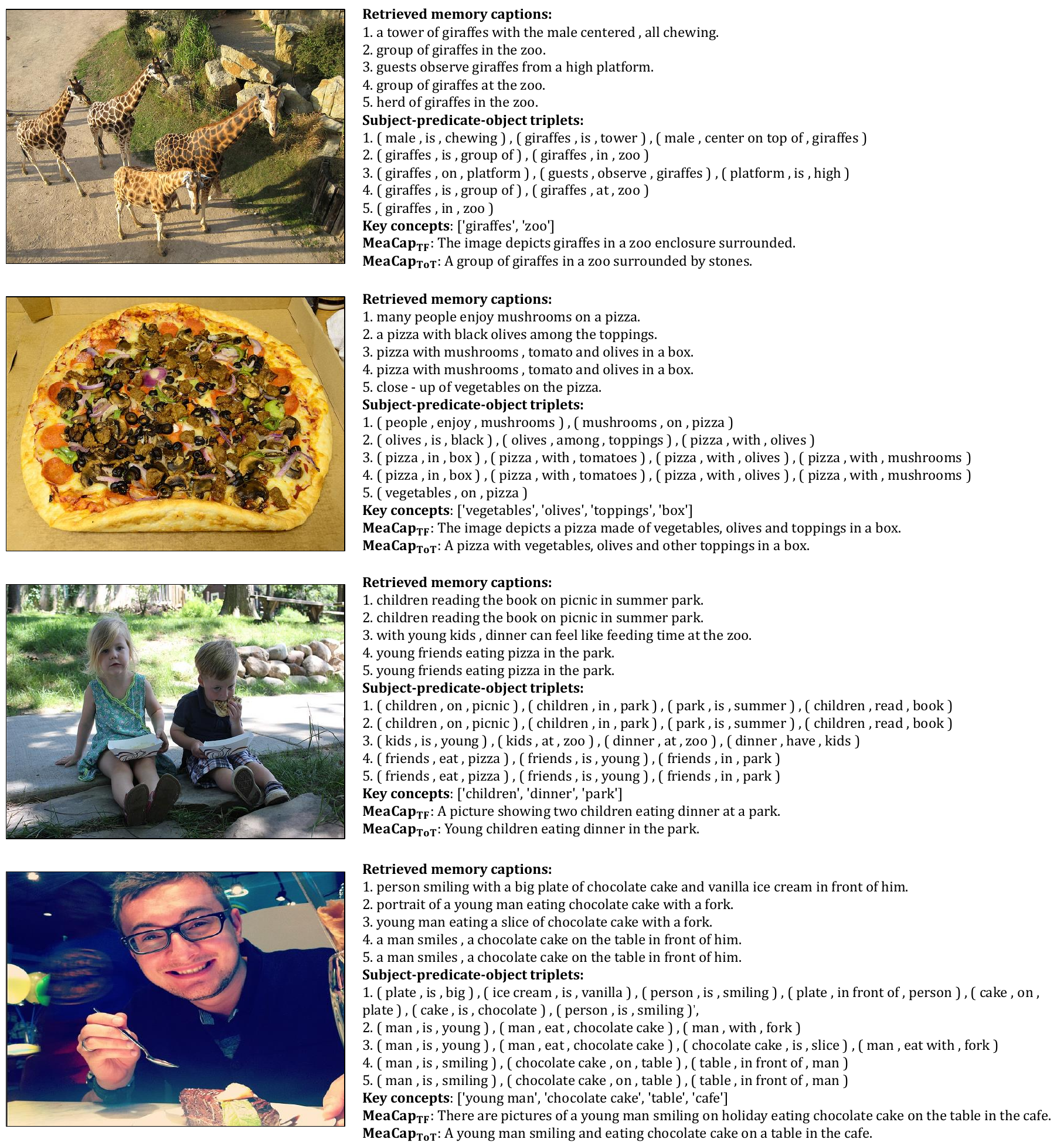}
        \vspace{-2mm}
	\caption{More qualitative results.}\label{fig:qualitatives}
\end{figure*}

\begin{figure*}[!tb]
	\centering 
	\includegraphics[width=0.95\textwidth]{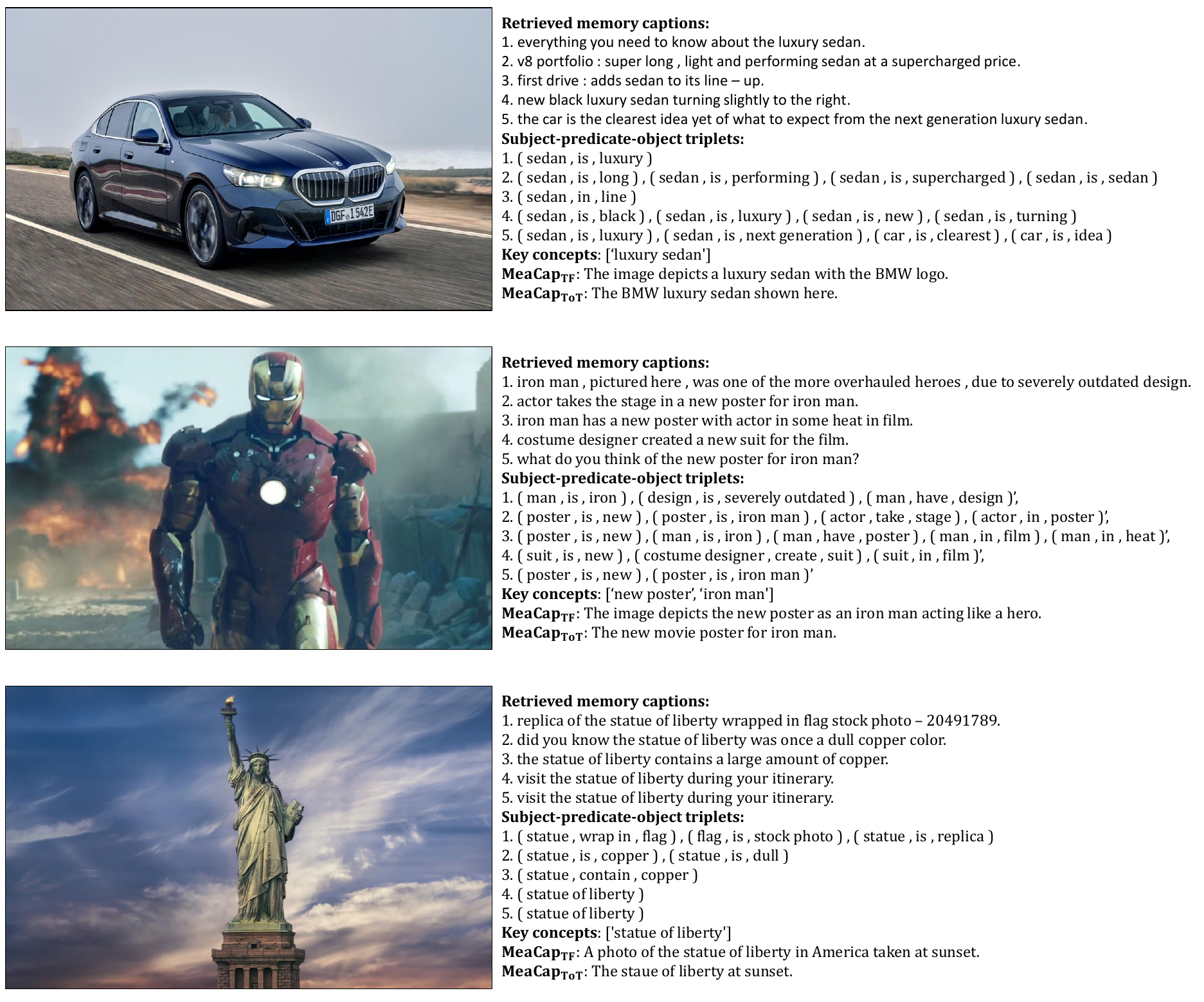}
        \vspace{-2mm}
	\caption{More qualitative results with world knowledge.}\label{fig:qualitatives_world}
\end{figure*}

Fig.~\ref{fig:qualitatives} and Fig.~\ref{fig:qualitatives_world} have shown some generated results with our proposed methods.
For each image, we have shown the intermediate results of the retrieve-then-filter module, $\ie$ the retrieved memory captions, the extracted subject-predicate-object triplets, and the filtered key concepts.
As we can see, we effectively filter the correct key concepts out.
Based on the key concepts, MeaCap$_\mathrm{TF}$ and MeaCap$_\mathrm{ToT}$ can generate captions with high consistency with image content.
Fig.~\ref{fig:qualitatives_world} showcases that the extracted key concepts from memory contain much world knowledge and can also alleviate the knowledge-forgotten phenomenon of text-only-training methods.

% WARNING: do not forget to delete the supplementary pages from your submission 
% \input{sec/X_suppl}

\end{document}

%% file: preamble.tex
\usepackage[dvipsnames]{xcolor}